# EHRSummarizer: A Privacy-Aware, FHIR-Native Reference Architecture for Source-Grounded EHR Summarization

Houman Kazemzadeh [1,2,3], Nima Minaifar [1,2], Kamyar Naderi [1,2,3], Sho Tabibzadeh [4]

[1] MedLedger365, Dubai, United Arab Emirates
[2] MedConnect365, Dubai, United Arab Emirates
[3] Xylemed, Dubai, United Arab Emirates
[4] Kypath Associates Inc., Greater Toronto Area, Canada

### Scope Statement

This technical report is intentionally limited in scope. It describes a reference architecture and prototype behavior for EHR summarization rather than a validated clinical intervention, autonomous clinical decision-support system, or evidence of clinical benefit. All examples and demonstrations are based on synthetic or test FHIR environments unless otherwise stated. Any future clinical use would require local governance review, privacy and security assessment, and appropriate evaluation before reliance in care workflows.

## Abstract

Clinicians routinely navigate fragmented electronic health record (EHR) interfaces to assemble a coherent picture of a patient's problems, medications, recent encounters, and longitudinal trends. This manuscript describes EHRSummarizer, a privacy-aware, FHIR-native reference architecture for structured EHR summarization. The architecture retrieves a targeted set of high-yield HL7 FHIR R4 resources, normalizes them into a clinical context package, and uses a constrained summarization stage to produce source-grounded summaries intended to support chart review. The architecture further clarifies missing-data status handling, medication-status ambiguity, controlled use of narrative clinical documents when available, and future source-to-summary traceability. The manuscript describes a reference architecture and prototype behavior rather than a validated clinical intervention, autonomous clinical decision-support system, or evidence of clinical benefit. Prototype demonstrations on synthetic and test FHIR environments illustrate end-to-end behavior and output formats; however, this manuscript does not report clinical outcomes, controlled workflow studies, or benchmark results. We outline an evaluation plan centered on faithfulness, omission risk, temporal correctness, usability, privacy, and operational monitoring to guide future institutional assessment.

## Keywords

## 1. Introduction

EHR platforms have increased the availability of patient data, but they often require clinicians to traverse numerous screens, tabs, and documents to assemble a coherent clinical narrative. This fragmentation can slow decision-making in time-limited workflows and contribute to documentation burden and cognitive overload.

While modern EHRs provide dashboards and filters, they frequently optimize for data entry, billing, and resource-specific viewing rather than rapid synthesis of a patient's current status and longitudinal trajectory. As a result, chart review often becomes an exercise in manual information retrieval: identifying active problems, linking medications to indications, reconstructing timelines, and interpreting trends across laboratory and vital signs.

Recent advances in generative modeling have made it possible to explore the conversion of structured clinical data into concise, clinician-oriented summaries. However, EHR summarization in practice requires more than text generation. It requires careful resource selection, normalization, transparency about missing information, deployment choices that align with privacy requirements, and guardrails that limit unsupported inference.

This paper describes a FHIR-native reference architecture intended to support structured summarization while minimizing data retention and enabling deployment within a healthcare organization's trust boundary. The goal is not to provide diagnoses or treatment directives, but to assist chart review by presenting available information in a consistent and reviewable format.

This manuscript makes five architectural contributions:

1. A resource-targeted retrieval strategy for summarization that prioritizes commonly populated and routinely consulted FHIR R4 resources.
2. A normalization step that constructs a clinical context package as a stable intermediate representation between EHR retrieval and summarization.
3. A proposed missing-data status layer that distinguishes unavailable, empty, errored, and intentionally unqueried domains rather than silently implying absence.
4. A privacy-aware design that supports stateless processing, data minimization, and configurable trust-boundary placement without prioritizing any single deployment model.
5. A safety posture and evaluation blueprint emphasizing faithfulness, omission risk, temporal correctness, and operational monitoring rather than clinical outcome claims.

Beyond summarization, the architecture is designed to provide a consistent representation of clinical context across heterogeneous EHR implementations, supporting standardized chart review workflows independent of vendor-specific interfaces.

## 2. Problem Statement and Motivation

### 2.1 Clinical burden and cognitive overload

Clinicians often spend substantial time navigating EHR interfaces and searching across notes, laboratory values, vital signs, imaging reports, and medication lists to answer recurring clinical questions such as:

1. What are the key active and historical problems?
2. Which medications appear current, ordered, administered, dispensed, or discontinued?
3. What major procedures, admissions, and investigations have occurred recently?
4. What laboratory or vital-sign trends require clinician review?
5. Are allergies, critical flags, or clinically relevant omissions visible?

This overhead contributes to clinician workload and reduces time available for direct patient care. A well-bounded EHR summary may help organize available information, but it should not substitute for clinician review of the underlying record.

### 2.2 Interoperability exists; synthesis remains difficult

Even when institutions expose data via FHIR, clinical synthesis remains difficult because data are:

1. distributed across multiple resource types;
2. not consistently coded;
3. incomplete or inconsistently populated across vendors and sites;
4. represented differently across structured resources and narrative notes;
5. not always presented in a clinician-friendly summary form.

### 2.3 Design goals

We designed the architecture with four goals:

1. **Clinical usefulness:** produce summaries structured for clinician review across specialties.
2. **Interoperability-first:** use FHIR as the default interface; if a source is non-FHIR, map it into a FHIR-aligned internal structure where feasible.
3. **Privacy-aware approach:** minimize retention and support deployments that keep sensitive data within the healthcare organization's governance boundary.
4. **Evidence discipline:** expose uncertainty, avoid unsupported inference, and make clear that evaluation is required before clinical reliance.

## 3. System Overview

All architectural constraints described in this manuscript are intended to be enforced outside the summarization model itself and to remain invariant under substitution of the underlying generative component. The summarization component may be implemented using a large language model or another text-generation method; however, the architectural properties described here are intended to remain independent of the specific model class.

### 3.1 High-level pipeline

The architecture contains five main stages:

1. **EHR retrieval:** query a FHIR endpoint for a targeted set of resources for a given patient.
2. **Normalization and structuring:** convert returned resources into a consistent internal schema, preserve timestamps, and reduce redundancy.
3. **Context packaging:** compile normalized elements into a summary-ready clinical context package with explicit domain status information.
4. **Constrained summarization:** generate a structured narrative from the context package without adding diagnoses, treatment recommendations, or unsupported facts.
5. **Presentation:** render the output in a clinician-facing review format. Optional follow-up questions, if enabled for research or pilot assessment, remain grounded in the same context package.

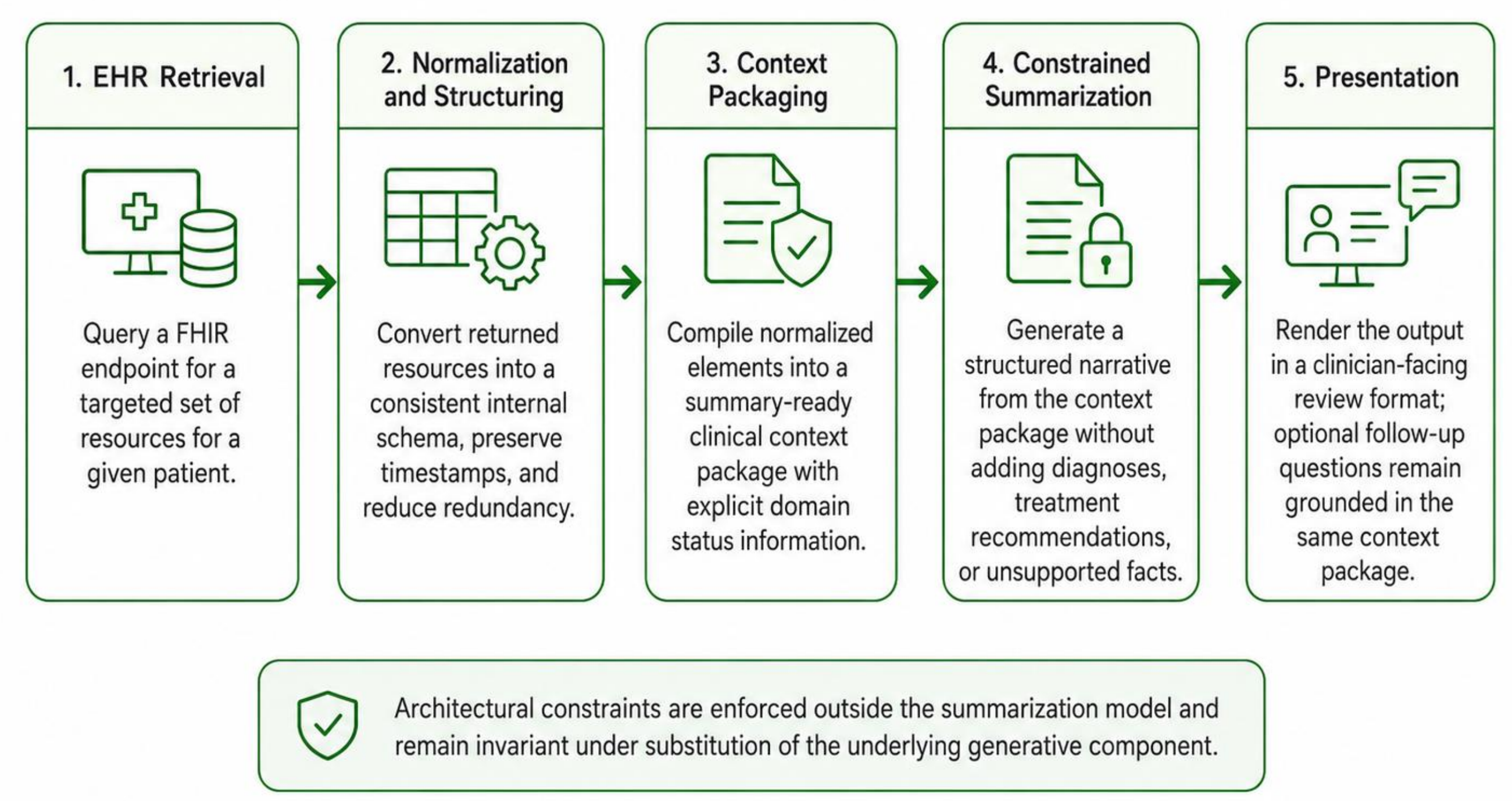

**Figure 1. High-level EHRSummarizer pipeline.** The architecture separates EHR retrieval, normalization and structuring, context packaging, constrained summarization, and presentation. Architectural constraints are applied outside the summarization model, allowing the retrieval and context-packaging layers to remain independent of the underlying generative component.

### 3.2 Why a resource-targeted approach

Rather than ingesting the entire chart indiscriminately, the architecture focuses on resource types that generally carry value for routine chart review: demographics, conditions, observations, procedures, medication-related resources, encounters, reports, documents, allergies, and flags. In this context, "high-yield" refers to resource types that are commonly populated across FHIR implementations and frequently consulted during chart review, not to automated clinical prioritization.

## 4. Data Sources and Resource Selection

### 4.1 Data sources

The architecture is designed to operate on EHR data exposed through HL7 FHIR R4-compatible interfaces. In development and prototype demonstrations, end-to-end behavior should be evaluated using test or synthetic FHIR environments and representative patient records to ensure repeatability without introducing real patient identifiers into engineering workflows. The same retrieval and summarization logic may be adapted to live clinical systems only subject to site configuration, authorization, governance controls, and institutional assessment.

### 4.2 Core FHIR resources considered

The architecture prioritizes resources that are commonly implemented across FHIR servers and relevant to structured clinical summarization. Availability is expected to vary by site. The retrieval layer may query the following resources when available:

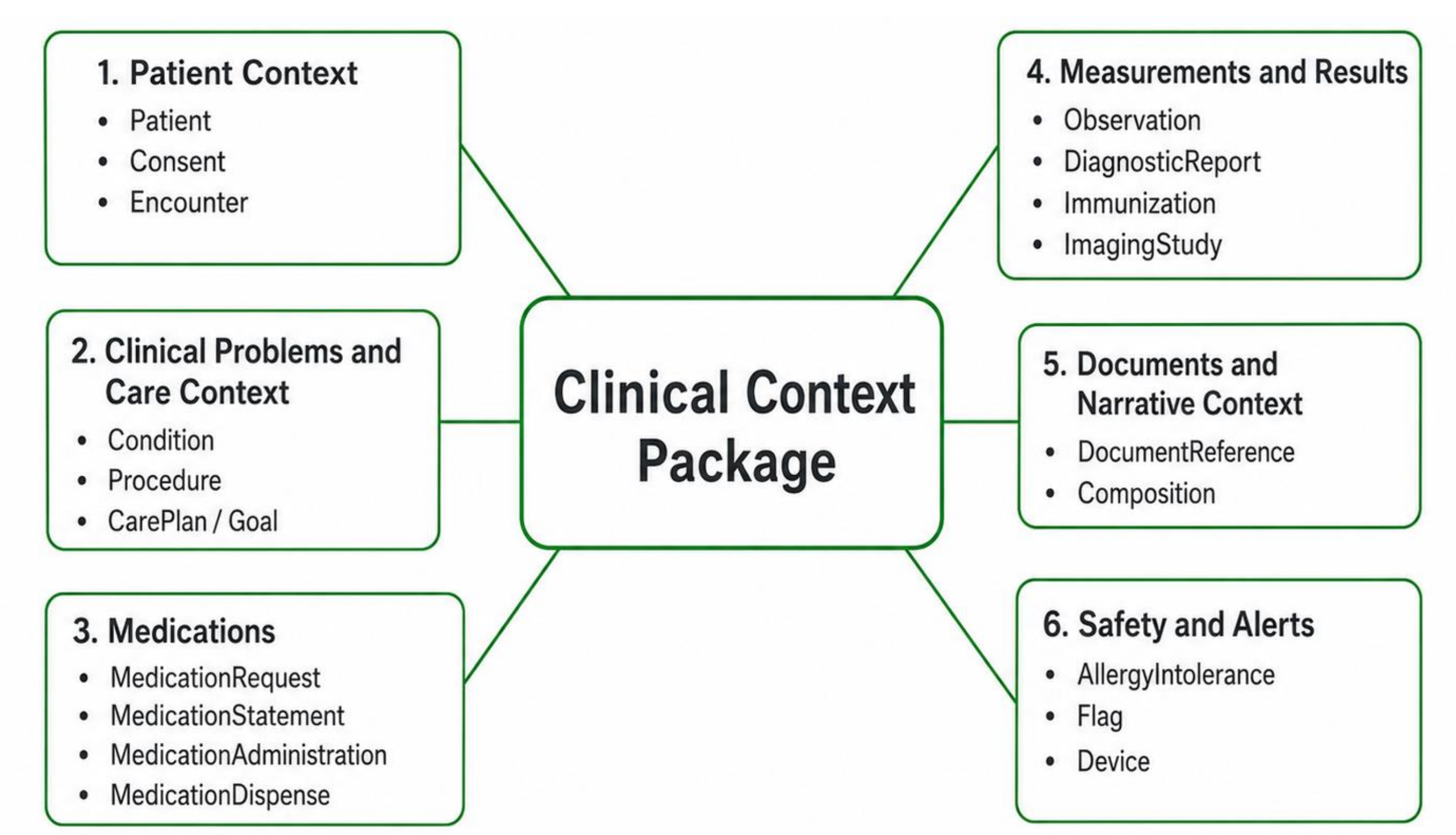

**Figure 2. Conceptual grouping of FHIR resources used in the clinical context package.** Resources are organized into patient context, clinical problems and care context, medications, measurements and results, documents and narrative context, and safety and alerts. Availability and coding depth are expected to vary across FHIR implementations; therefore, resource groups should be interpreted together with domain-status reporting and local governance constraints.

**Table 1.** Core FHIR resources considered for structured EHR summarization.

| *FHIR resource* | Role in EHR summary | Caution |
|---|---|---|
| *Patient* | Demographics and core identifiers that anchor the summary. | Interpret cautiously; local coding varies. |
| *Consent* | Consent status and constraints relevant to access and downstream processing, when exposed. | Optional / site-dependent. |
| *Condition* | Active and historical problems or diagnoses, recognizing that problem lists may be outdated or incomplete. | Interpret cautiously; local coding varies. |
| *Observation* | Vital signs, laboratory values, and selected social-history markers depending on coding practices. | Interpret cautiously; local coding varies. |
| *MedicationRequest* | Medication orders and prescribing intent; not sufficient by itself to prove active patient use. | Interpret cautiously; local coding varies. |
| *MedicationStatement / MedicationAdministration / MedicationDispense* | Optional medication-status resources where available; useful for distinguishing reported use, administered medications, and dispensed medications. | Optional / site-dependent. |
| *Procedure* | Surgeries, interventions, and performed clinical actions. | Optional / site-dependent. |
| *Encounter* | Visit context, timestamps, and continuity across care episodes. | Interpret cautiously; local coding varies. |
| *DiagnosticReport* | Diagnostic conclusions and structured panels, including selected report context depending on server content. | Optional / site-dependent. |
| *DocumentReference / Composition* | Controlled access to narrative documents, discharge-related documents, reports, or structured compositions when source boundaries can be preserved. | Optional / site-dependent. |
| *AllergyIntolerance* | Allergies and adverse reactions; should receive explicit status reporting rather than silent omission. | Optional / site-dependent. |
| *Immunization* | Vaccination history and preventive care signals. | Optional / site-dependent. |
| *CarePlan / Goal* | Care coordination artifacts and longitudinal targets where available. | Optional / site-dependent. |
| *ImagingStudy* | Imaging metadata, not image pixels, to contextualize radiology workflows while images remain in PACS. | Optional / site-dependent. |
| *Flag / Device* | Alerts, warnings, and implanted or used devices where properly encoded. | Optional / site-dependent. |

Vendor variability is expected. Many real-world FHIR servers omit some resources, partially encode data, or differ in which patient-facing references they support. Persistent variability in FHIR completeness and coding practices is treated as a foundational constraint informing the design of vendor-agnostic synthesis.

### 4.3 Retrieval strategy and clinical context package

Rather than retrieving the entire longitudinal record, the architecture retrieves a targeted set of high-yield resources and compiles them into a compact intermediate representation, the clinical context package. This package is a normalized summary-ready structure that:

1. groups content by clinical topic;
2. preserves timestamps where available;
3. minimizes redundancy, such as repeated medication orders or duplicated procedure entries;
4. records the status of each domain, including whether it was queried, unavailable, empty, or errored;
5. produces stable section headers that map cleanly into downstream formatting.

This intermediate step is important because raw FHIR bundles are heterogeneous, verbose, and often contain repeated or partially redundant information across encounters.

## 4.4 Proposed handling of absent or uncertain domains

A key refinement is to distinguish absence of evidence from evidence of absence. Instead of silently omitting every empty domain, the clinical context package can attach a simple status label to each clinical domain. This is a proposed architectural improvement intended to reduce ambiguity in future implementations.

**Table 2.** Proposed domain-status labels for absent or uncertain EHR data.

| ***Domain status*** | **Meaning** |
|---|---|
| *Present* | Data were retrieved and summarized. |
| *Queried-empty* | The domain was queried and returned no entries. |
| *Unavailable* | The FHIR endpoint does not expose the resource, or the resource is not supported at the site. |
| *Error* | The resource was expected, but retrieval failed or timed out. |
| *Not queried* | The domain was outside the current retrieval configuration. |

## 5. Summarization Stage: Hosted and Local Execution Options

The summarization stage supports more than one execution pattern. This manuscript does not prioritize a specific deployment model; rather, it separates retrieval, normalization, context packaging, and summarization so that institutions can evaluate options against their own privacy, security, operational, and governance requirements.

## 5.1 Model execution options

A hosted model endpoint may receive the clinical context package and return a structured clinician-facing summary. This approach may reduce local infrastructure requirements and operational responsibility, but requires robust contractual, privacy, security, and network controls.

A locally deployed model may run within an organization's controlled environment, including on-premises or private-cloud infrastructure. This pattern may keep processing within the organization's infrastructure boundary, but increases operational burden, including model packaging, monitoring, upgrades, and performance tuning.

The interface between the context package and summary is intentionally consistent, allowing hosted and local inference options to be evaluated without reworking the retrieval pipeline or summary schema.

## 5.2 Prompting and guardrails

The summarizer is instructed to:

1. Produce a concise clinical summary organized into predefined sections;
2. Avoid repeating the input context and avoid conversational filler;
3. Summarize only information present in the clinical context package;
4. State unavailable or uncertain domains when clinically relevant and supported by the context package status layer;
5. Avoid diagnostic or treatment recommendations;
6. In optional interactive mode, restrict follow-up behavior to clarification, navigation, and summarization grounded in the same context package.

Future iterations may add statement-level traceability by linking summary claims to supporting elements in the context package. This is proposed as an auditability feature and is not reported here as an empirically validated capability.

## 6. Output Formats: Text-Based Summaries and Structured Views

### 6.1 Text output

The primary output is a clinician-readable EHR summary organized into clear sections. This format is straightforward to render, archive, and review. It should be interpreted as a compressed representation of available data rather than as a substitute for source records.

### 6.2 Structured/table output

In addition to narrative text, the architecture can support table-aligned rendering in which each section is displayed as a structured block, such as conditions, medications, procedures, encounters, allergies, or laboratory trends. Such rendering may improve readability and review consistency, but should preserve timestamps, source boundaries, and uncertainty markers.

### 6.3 Optional clinician interaction

An optional interaction panel may allow clinicians to ask follow-up questions grounded in the same context package or generated summary. Because such interaction can increase risk if not constrained, it should be treated as optional, site-governed, and disableable. Responses should not propose diagnoses or treatment plans and should direct the clinician back to source records when uncertainty exists.

## 7. Practical Considerations for Deployment Alignment

### 7.1 Stateless processing and data minimization

The architecture can be configured for stateless operation, where context packages are processed transiently and discarded after summary generation. Where retention is required for auditability, storage can be limited to the summary artifact, retrieval status metadata, and minimal operational logs rather than raw upstream EHR payloads.

### 7.2 Reliability and failure handling

The retrieval layer is expected to encounter missing resources, partial records, inconsistent coding, pagination, and intermittent server errors. The pipeline should report such conditions in clinician-readable form. For example, reporting “AllergyIntolerance resource unavailable” is more transparent than silent omission because it avoids implying that no allergies exist.

## 8. Privacy, Security, Governance, and Scope Boundaries

### 8.1 Privacy-aware processing and data minimization

The architecture is designed to support privacy-aware clinical workflows by limiting the amount of EHR data transmitted, processed, and retained. Only a targeted subset of resources is retrieved per patient, and only elements required for summarization are maintained in the normalized clinical context package.

### 8.2 Deployment patterns and trust boundaries

The architecture supports hosted inference and local inference as two possible trust-boundary patterns. Hosted inference may reduce local operational responsibility but requires contractual, privacy, security, and network safeguards. Local inference keeps processing inside the healthcare organization's controlled environment but increases operational responsibility. This manuscript does not claim superiority of either pattern.

### 8.3 Authorization and patient awareness

In any future deployment, patient access and clinician access should be governed by the site's identity and authorization mechanisms, such as SMART on FHIR or OAuth2-aligned flows where applicable. Patient awareness of access, if required by institutional policy, may be implemented through existing audit logs, portal notifications, or consent workflows.

### 8.4 Security considerations

A practical implementation should consider role-based access control, encryption in transit and at rest when retention is enabled, audit logging of access and summary-generation events, rate limiting, safeguards against bulk extraction, environment segmentation, least privilege, and secret management. These are architectural considerations rather than claims of completed certification.

### 8.5 Threat model and failure modes

Key threats include exposure of protected health information outside the intended trust boundary, unsupported or inferred clinical statements, omission of clinically relevant elements that are present in the source data, temporal errors that misrepresent trends or encounter chronology, and overreliance on a generated summary. Mitigations include data minimization, stateless processing where feasible, strict grounding to the context package, explicit missing-data reporting, and pre-deployment testing using adversarial and longitudinal cases.

### 8.6 Governance boundary

This manuscript frames EHRSummarizer as a reference architecture for EHR summarization and chart-review support, not as autonomous clinical decision support. Future implementations should be assessed locally before clinical deployment, especially if the scope changes from summarization of available facts toward recommendation, diagnosis, triage, or treatment guidance.

## 9. Evaluation Plan

Because this manuscript does not report benchmark results or clinical workflow studies, evaluation is framed as a future requirement. The plan below focuses on measurements that can be performed using the retrieved context package as a reference source of truth.

### 9.1 Clinical coverage and correctness

**Objective:** assess whether the generated EHR summary captures clinically relevant content present in the retrieved data without adding unsupported statements.
**Suggested approach:** create clinician-defined checklists for demographics, active problems, major historical problems, medication status, allergies, key recent laboratory values and vital signs, major procedures, encounter context, and missing-data status. Compare summary output against the context package.

**Candidate measures:**

1. Section-level completeness score.
2. Faithfulness and contradiction rate.
3. Omission risk for clinically relevant domains.
4. Error categorization: omission, incorrect value, temporal error, hallucination, and unsupported inference.

### 9.2 Temporal correctness

**Objective:** assess whether dates, encounter chronology, recent-versus-historical distinctions, and laboratory trajectories are preserved.
**Suggested approach:** include longitudinal cases with repeated measurements, discontinued medications, duplicate orders, old diagnoses, and multiple admissions.

### 9.3 Time-to-information and workflow assessment

**Objective:** assess whether use of the system is associated with changes in the time required to locate and synthesize key patient information during common workflows.
**Suggested approach:** compare standard chart review with summary-assisted review using representative scenarios, while avoiding claims of benefit until controlled results are available.

**Candidate measures:**

1. Time to answer scenario-specific questions.
2. Number of EHR screens visited or interaction steps.
3. Clinician-rated cognitive workload.
4. Trust, perceived usefulness, and need for source verification.

### 9.4 Usability and adoption

**Objective:** assess whether the summary format is readable, appropriately scoped, and compatible with clinical documentation habits.
**Suggested approach:** mixed-method evaluation may include surveys, interviews, and structured feedback on section relevance, terminology, missing-data display, and desired navigation to source records.

## 9.5 Safety testing and failure-mode analysis

**Objective:** identify and characterize potentially unsafe behaviors before any clinical use.
**Suggested approach:** stress tests should include missing resources, conflicting observations, duplicate medication orders, highly longitudinal lab histories, ambiguous narrative documents, and source documents with outdated or contradictory information.

## 10. Prototype Demonstrations: Non-controlled Observations

Prototype demonstrations in synthetic and test FHIR environments can exercise end-to-end system behavior from retrieval to normalization to summary rendering and can surface practical failure modes such as missing resources, duplicated medication records, inconsistent coding, and incomplete narrative context. These demonstrations are not controlled clinical studies and should not be interpreted as evidence of improved outcomes, safety, time savings, or workflow effectiveness. Their purpose is to inform subsequent institutional evaluation and the monitoring practices described above.

## 11. Limitations

This work describes a reference architecture and prototype behavior rather than a validated clinical intervention. The manuscript does not report outcomes from real-world patient care, and prototype demonstrations are limited to synthetic and test FHIR environments. Summary quality may vary with local coding practices, incomplete FHIR coverage, resource availability, and institutional differences in documentation.

Several limitations should be considered:

1. **Data quality and coding variability:** summaries can only be as complete and correct as the upstream data.
2. **Resource availability differences:** not all systems expose the same FHIR resources or linkage patterns.
3. **Narrative context:** important clinical nuance may exist only in unstructured notes or external documents not represented in retrieved resources.
4. **Medication ambiguity:** MedicationRequest alone may represent orders rather than actual current use; future designs should distinguish ordered, dispensed, administered, reported, and discontinued medications when resources permit.
5. **Generative model constraints:** even grounded summaries can compress, omit, or misorder information. Clinician review remains essential.
6. **Scope boundaries:** this manuscript avoids diagnostic, treatment, or autonomous decision-making claims, and future use would require local governance and evaluation.
7. **No benchmark or clinical study:** controlled evaluation is required before reliance in clinical workflows.

## 12. Future Work

### 12.1 Source-to-summary traceability

Future work may explore statement-level traceability, in which summary statements link back to the context-package elements that support them. Such traceability could support auditability, clinician review, and error analysis.

### 12.2 Controlled handling of unstructured documents

When DocumentReference or Composition resources include narrative reports, discharge-related documents, or scanned documents, future work may explore controlled extraction methods that preserve source boundaries, document date, document type, authorship where available, and uncertainty. Narrative content should not be blended into structured sections without provenance.

### 12.3 Medication-status reconciliation

Future work may explore methods for distinguishing ordered medications from administered, dispensed, reported, and discontinued medications when FHIR resources permit. If medication status cannot be determined, the summary should indicate uncertainty rather than presenting an ambiguous list as current therapy.

### 12.4 Longitudinal trend summarization

Future work may enhance representation of trends, such as HbA1c, creatinine, hemoglobin, blood pressure, or weight, with clear time anchors and most-recent-versus-prior comparisons.

### 12.5 Specialty-aware templates

Configurable specialty emphasis may be useful for cardiology, oncology, radiology, emergency medicine, and primary care, but should preserve a standard baseline structure and source-grounding constraints.

### 12.6 Minimal synthetic benchmark

A future public or semi-public synthetic benchmark could support reproducible evaluation without exposing patient data. Such a benchmark could include edge cases involving missing resources, outdated problem lists, duplicate medication records, conflicting observations, and narrative-document ambiguity.

## 13. Conclusion

This work presents a privacy-aware, FHIR-native reference architecture for structured EHR summarization intended to support chart review. The approach emphasizes targeted retrieval of commonly used FHIR resources, normalization into a clinical context package, explicit handling of missing or uncertain domains, and configurable deployment within organizational governance boundaries. By constraining summarization to available evidence and avoiding diagnostic or treatment recommendations, the architecture is intended to mitigate the risk of unsupported inference while assisting clinician orientation. Rather than positioning EHR summarization as a predictive or decision-making capability, this manuscript frames it as an information-synthesis function that requires evaluation before clinical reliance. Future work should focus on controlled institutional evaluations to quantify faithfulness, omission risk, temporal correctness, usability, and workflow integration, as well as on operational monitoring to support responsible assessment of clinical summarization technologies.

## Acknowledgments

The authors thank internal clinical and engineering collaborators who provided feedback on summary structure, usability, and safety requirements.

## Funding

This work was funded and supported by MedConnect365, which provided organizational resources for system design, implementation, and internal evaluation. The manuscript is presented as a technical and architectural report and does not claim clinical efficacy, patient outcomes, or regulatory clearance.

## Competing Interests

The authors are employees, contractors, or collaborators of organizations involved in healthcare software, interoperability, and clinical summarization technologies, including MedConnect365 and related entities. EHRSummarizer is related to commercial EHR summarization technology under development. This manuscript describes a reference architecture and prototype behavior; it does not present evidence of clinical efficacy, validated clinical intervention, regulatory clearance, or patient outcomes.

## Data Availability

No patient-identifiable data are shared in this preprint. Demonstrations were conducted using synthetic or test FHIR environments and representative records.

## Code Availability

Code is not publicly available in this version due to commercialization and security considerations. To support reproducibility without releasing proprietary or security-sensitive components, the manuscript describes the data flow, interfaces, and proposed evaluation approach in detail.

## Ethics Statement

This work did not involve human subjects research; no identifiable patient data were used. Prototype demonstrations were performed in synthetic and test FHIR environments.